\documentclass[11pt,a4paper]{article}
\usepackage[hyperref]{acl2018}
\usepackage{CJKutf8}
\usepackage{times}
\usepackage{latexsym}
\usepackage{amsthm}
\usepackage{mathtools,amssymb}
\usepackage{booktabs}
\usepackage{multirow}
\usepackage{multicol}
\usepackage{array}
\usepackage{url}
 
\aclfinalcopy % Uncomment this line for the final submission
\def\aclpaperid{1279} %  Enter the acl Paper ID here

\newcommand*{\affaddr}[1]{#1} % No op here. Customize it for different styles.
\newcommand*{\affmark}[1][*]{\textsuperscript{#1}}

\title{Incorporating Chinese Characters of Words \\ for Lexical Sememe Prediction}
   
\author{\parbox{\linewidth}{\centering
Huiming Jin{\rm\affmark[1]}\thanks{Work done while doing internship at Tsinghua University.}\hspace{5pt}$^{\dag}$, Hao Zhu{\rm\affmark[2]}\thanks{\makebox[0.97\textwidth][s]{Equal contribution. Huiming Jin proposed the overall idea, designed the first experiment, conducted both experiments, and}\newline \makebox[0.97\textwidth][s]{wrote the paper; Hao Zhu made suggestions on ensembling, proposed the second experiment, and spent a lot of time on}\newline \mbox{proofreading the paper and making revisions. All authors helped shape the research, analysis and manuscript.}}\space, Zhiyuan Liu{\rm\affmark[2,3]}\thanks{Corresponding author: Z. Liu (liuzy@tsinghua.edu.cn)}\space, Ruobing Xie{\rm\affmark[4]}, \\ Maosong Sun{\rm\affmark[2,3]}, Fen Lin{\rm\affmark[4]}, Leyu Lin{\rm\affmark[4]}} \\
\affaddr{\affmark[1] Shenyuan Honors College, Beihang University, Beijing, China} \\
\affaddr{\affmark[2]  Beijing National Research Center for Information Science and Technology, \\ State Key Laboratory of Intelligent Technology and Systems, \\
Department of Computer Science and Technology, Tsinghua University, Beijing, China} \\
\affaddr{\affmark[3]Jiangsu Collaborative Innovation Center for Language Ability, \\
Jiangsu Normal University, Xuzhou 221009 China} \\
\affaddr{\affmark[4] Search Product Center, WeChat Search Application Department, Tencent, China}}
\date{}
 
\usepackage{scrextend}
\deffootnote[1.5em]{1.5em}{1em}{\thefootnotemark\space}

\begin{document}
\maketitle 
\begin{abstract}
Sememes are minimum semantic units of concepts in human languages, such that 
each word sense is composed of one or multiple sememes. Words are usually 
manually annotated with their sememes by linguists, and form linguistic 
common-sense knowledge bases widely used in various NLP tasks. Recently, the 
lexical sememe prediction task has been introduced. It consists of automatically 
recommending sememes for words, which is expected to improve annotation 
efficiency and consistency. However, existing methods of lexical sememe 
prediction typically rely on the external context of words to represent the 
meaning, which usually fails to deal with low-frequency and out-of-vocabulary 
words. To address this issue for Chinese, we propose a novel framework to take 
advantage of both internal character information and external context 
information of words. We experiment on HowNet, a Chinese sememe knowledge base, 
and demonstrate that our framework outperforms state-of-the-art baselines by a 
large margin, and maintains a robust performance even for low-frequency words. 
\footnote{\mbox{Code is available at \url{https://github.com/thunlp/Character-enhanced-Sememe-Prediction}}}

%By case study, we also demonstrate the effectiveness of our 
%framework for Chinese morphology analysis.    
%Sememes are minimum semantic units of concepts in human languages, such that each word 
%sense is composed of one or multiple sememes. 
%Words are usually manually annotated with their sememes by linguists,
%The sememes of word senses are 
%typically manually annotated by linguists
%and form linguistic common-sense 
%knowledge bases widely used in various NLP tasks. Recently, the lexical sememe 
%prediction task is explored to automatically recommend sememes for words, which 
%is expected to improve annotation efficiency and consistency. However, existing 
%methods of lexical sememe prediction typically rely on external context of words 
%to represent word meanings for prediction, which usually fail to deal with those 
%low-frequency and out-of-vocabulary words. To address this issue, we take 
%Chinese as an example and propose a novel framework to take advantages of both 
%internal character information and external context information of a word for 
%sememe prediction. We conduct experiments on HowNet, a Chinese sememe knowledge 
%base, and the results demonstrate that our framework outperforms the 
%state-of-the-art baselines by a large margin and stays robust performance in the 
%long-tail scenario. By case study, we also demonstrate the effectiveness of our 
%framework for Chinese morphology analysis.
\end{abstract}
 
\section{Introduction}
A sememe is an indivisible semantic unit for human languages defined by
linguists \cite{bloomfield1926set}. The semantic meanings of concepts (e.g.,
words) can be composed by a finite number of sememes. However, the sememe set of
a word is not explicit, which is why linguists build knowledge bases (KBs) to 
annotate words with sememes manually.

\begin{CJK*}{UTF8}{gbsn}
  \begin{figure}
    \centering
    \includegraphics[width=0.965\linewidth]{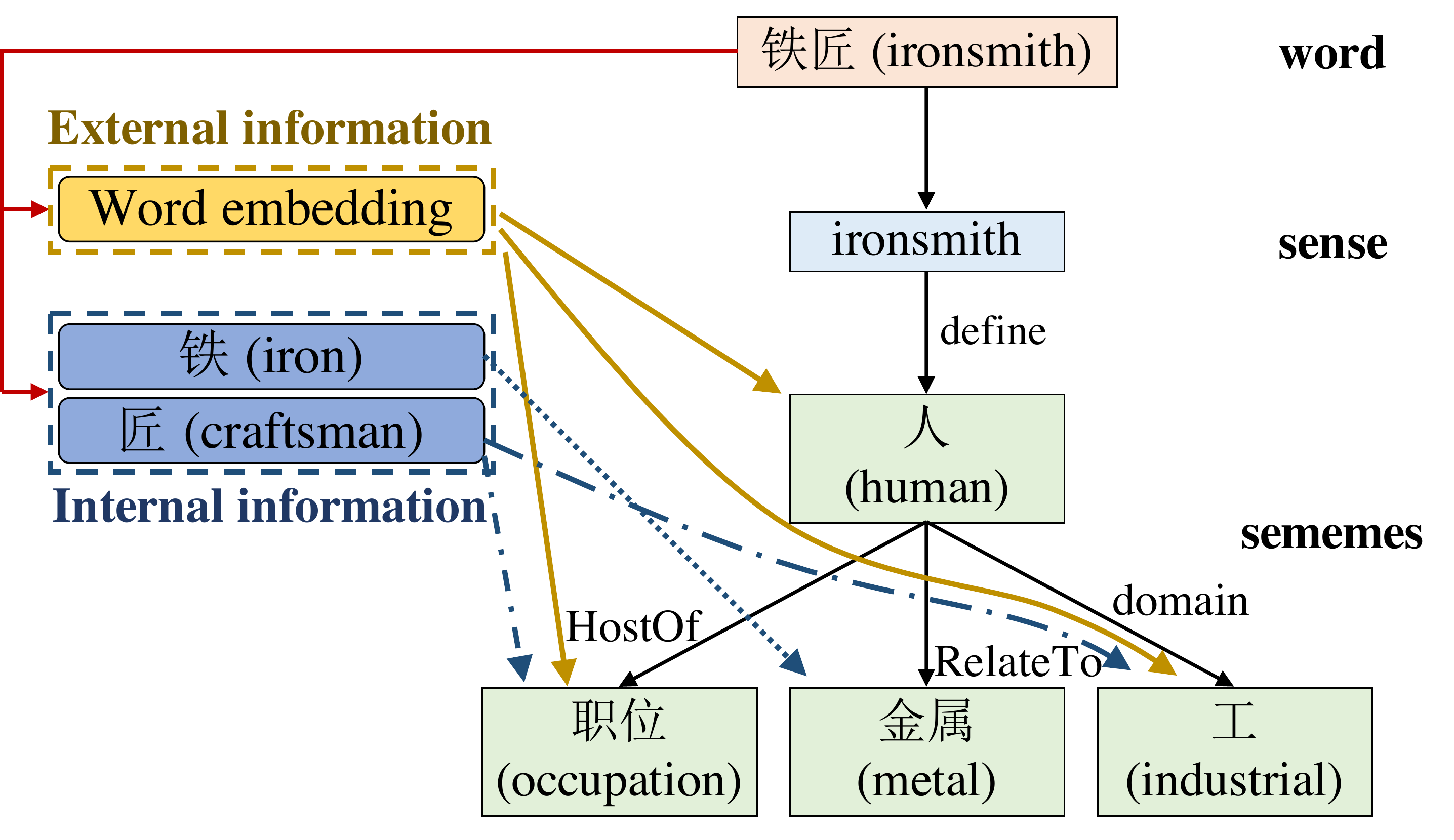}
    \caption{Sememes of the word ``铁匠'' (ironsmith) in HowNet, where 
      \emph{occupation}, \emph{human} and \emph{industrial} can be inferred by 
      both external (contexts) and internal (characters) information, while 
      \emph{metal} is well-captured only by the internal information within the 
      character `` 铁'' (iron).}
    \label{fig:ins-blacksmith}
  \end{figure}
\end{CJK*}

HowNet is a classical widely-used sememe KB \cite{zhendong2006hownet}. In 
HowNet, linguists manually define approximately $2,000$ sememes, and annotate 
more than $100,000$ common words in Chinese and English with their relevant 
sememes in hierarchical structures. HowNet is well developed and has a wide 
range of applications in many NLP tasks, such as word sense 
disambiguation~\cite{duan2007word}, sentiment 
analysis~\cite{xianghua2013multi,huang2014new} and cross-lingual word 
similarity~\cite{xia2011measuring}.

Since new words and phrases are emerging every day and the semantic meanings of
existing concepts keep changing, it is time-consuming and work-intensive for 
human experts to annotate new \newpage \noindent concepts and maintain 
consistency for large-scale sememe KBs. To address this issue, 
\newcite{xie2017lexical} propose an automatic sememe prediction framework to 
assist linguist annotation. They assumed that words which have similar semantic 
meanings are likely to share similar sememes. Thus, they propose to represent 
word meanings as embeddings \cite{pennington2014glove,mikolov2013distributed} 
learned from a large-scale text corpus, and they adopt collaborative filtering 
\cite{sarwar2001item} and matrix factorization \cite{koren2009matrix} for sememe 
prediction, which are concluded as Sememe Prediction with Word Embeddings (SPWE) 
and Sememe Prediction with Sememe Embeddings (SPSE) respectively. However, those 
methods ignore the internal information within words (e.g., the characters in 
Chinese words), which is also significant for word understanding, especially for
words which are of low-frequency or do not appear in the corpus at all. In this 
paper, we take Chinese as an example and explore methods of taking full 
advantage of both external and internal information of words for sememe 
prediction. 

\begin{CJK*}{UTF8}{gbsn}
In Chinese, words are composed of one or multiple characters, and most
characters have corresponding semantic meanings. As shown by
\newcite{yin298quantitative}, more than $90\%$ of Chinese characters in modern
Chinese corpora are morphemes. Chinese words can be divided into single-morpheme
words and compound words, where compound words account for a dominant
proportion. The meanings of compound words are closely related to their internal
characters as shown in Fig.\ \ref{fig:ins-blacksmith}. Taking a compound word 
``铁匠'' (ironsmith) for instance, it consists of two Chinese characters: 
``铁'' (iron) and ``匠'' (craftsman), and the semantic meaning of ``铁匠'' can 
be inferred from the combination of its two characters ({\it iron} + 
{\it craftsman} $\,\to\,$ {\it ironsmith}). Even for some single-morpheme words, 
their semantic meanings may also be deduced from their characters. For 
example, both characters of the single-morpheme word ``徘徊'' (hover) represent 
the meaning of ``hover'' or ``linger''. Therefore, it is intuitive to take the 
internal character information into consideration for sememe prediction.
%Additionally, there are only about $4,200$ characters that take up the top
%$99.94\%$ most popular words \cite{yin298quantitative}, which means
%incorporating the internal information could be a good solution for the
%out-of-vocabulary problem.
\end{CJK*} 

\begin{CJK*}{UTF8}{gbsn}
In this paper, we propose a novel framework for Character-enhanced Sememe 
Prediction (CSP), which leverages both internal character information and 
external context for sememe prediction. CSP predicts the sememe candidates for a 
target word from its word embedding and the corresponding character embeddings. 
Specifically, we follow SPWE and SPSE as introduced by \newcite{xie2017lexical} 
to model external information and propose Sememe Prediction with 
Word-to-Character Filtering (SPWCF) and Sememe Prediction with Character and 
Sememe Embeddings (SPCSE) to model internal character information. In our 
experiments, we evaluate our models on the task of sememe prediction using 
HowNet. The results show that CSP achieves state-of-the-art performance and 
stays robust for low-frequency words.

To summarize, the key contributions of this work are as follows: (1) To the best 
of our knowledge, this work is the first to consider the internal information of
characters for sememe prediction. (2) We propose a sememe prediction framework 
considering both external and internal information, and show the effectiveness
and robustness of our models on a real-world dataset.
\end{CJK*}

\section{Related Work}
\paragraph{Knowledge Bases.}
Knowledge Bases (KBs), aiming to organize human knowledge in structural forms, 
are playing an increasingly important role as infrastructural facilities of 
artificial intelligence and natural language processing. KBs rely on manual 
efforts \cite{bollacker2008freebase}, automatic extraction \cite{auer2007dbpedia}, 
manual evaluation \cite{suchanek2007yago}, automatic completion and alignment 
\cite{bordes2013translating,toutanova2015representing, zhu2017iterative} to 
build, verify and enrich their contents. WordNet 
\cite{miller1995wordnet} and BabelNet \cite{navigli2012babelnet} are the 
representative of linguist KBs, where words of similar meanings are grouped to 
form thesaurus \cite{nastase2001word}. Apart from other linguistic KBs, sememe 
KBs such as HowNet \cite{zhendong2006hownet} can play a significant role in 
understanding the semantic meanings of concepts in human languages and are 
favorable for various NLP tasks: information structure annotation 
\cite{gan2000annotating}, word sense disambiguation \cite{gan2002knowledge}, 
word representation learning \cite{niu2017improved,faruqui2015retrofitting}, and 
sentiment analysis \cite{xianghua2013multi} inter alia. Hence, lexical sememe 
prediction is an important task to construct sememe KBs.

\paragraph{Automatic Sememe Prediction.} Automatic sememe prediction is proposed 
by \newcite{xie2017lexical}. For this task, they propose SPWE and SPSE, which 
are inspired by collaborative filtering \cite{sarwar2001item} and matrix 
factorization \cite{koren2009matrix} respectively. SPWE recommends the sememes 
of those words that are close to the unlabelled word in the embedding space. 
SPSE learns sememe embeddings by matrix factorization \cite{koren2009matrix} 
within the same embedding space of words, and it then recommends the most 
relevant sememes to the unlabelled word in the embedding space. In these 
methods, word embeddings are learned based on external context information 
\cite{pennington2014glove,mikolov2013distributed} on large-scale text corpus. 
These methods do not exploit internal information of words, and fail to handle 
low-frequency words and out-of-vocabulary words. In this paper, we propose to 
incorporate internal information for lexical sememe prediction.

\paragraph{Subword and Character Level NLP.}  Subword and character level NLP 
models the internal information of words, which is especially useful to address 
the out-of-vocabulary (OOV) problem. Morphology is a typical research area of 
subword level NLP. Subword level NLP has also been widely considered in many NLP 
applications, such as keyword spotting \cite{narasimhan2014morphological}, 
parsing \cite{seeker2015graph}, machine translation \cite{dyer2010cdec}, speech 
recognition \cite{creutz2007analysis}, and paradigm completion \cite{sutskever2014sequence,bahdanau2014neural,cotterell2016sigmorphon,kann2017one,jin2017exploring}. 
Incorporating subword information for word embeddings \cite{bojanowski2017enriching,cotterell2016morphological,chen2015joint,wieting2016charagram,yin2016multi} 
facilitates modeling rare words and can improve the performance of several NLP 
tasks to which the embeddings are applied. Besides, people also consider 
character embeddings which have been utilized in Chinese word segmentation 
\cite{10.1007/978-3-319-12640-1_34}. 

The success of previous work verifies the feasibility of utilizing internal 
character information of words. We design our framework for lexical sememe 
prediction inspired by these methods.

\section{Background and Notation}

In this section, we first introduce the organization of {\it sememes}, 
{\it senses} and {\it words} in HowNet. Then we offer a formal definition of 
lexical sememe prediction and develop our notation.

\subsection{Sememes, Senses and Words in HowNet}

\begin{CJK*}{UTF8}{gbsn}
HowNet provides sememe annotations for Chinese words, where each word is 
represented as a hierarchical tree-like sememe structure. Specifically, a word 
in HowNet may have various {\it senses}, which respectively represent the 
semantic meanings of the word in the real world. Each {\it sense} is defined as 
a hierarchical structure of {\it sememes}. For instance, as shown in the right 
part of Fig.\ \ref{fig:ins-blacksmith}, the word ``铁匠'' (ironsmith) has one 
sense, namely {\it ironsmith}. The sense {\it ironsmith} is defined by the 
sememe ``人'' (human) which is modified by sememe ``职位'' (occupation), ``金属'' 
(metal) and ``工'' (industrial). In HowNet, linguists use about $2,000$ sememes 
to describe more than $100,000$ words and phrases in Chinese with various 
combinations and hierarchical structures.
\end{CJK*}

\subsection{Formalization of the Task}
In this paper, we focus on the relationships between the {\it words} and the 
{\it sememes}. Following the settings of \newcite{xie2017lexical}, we simply 
ignore the senses and the hierarchical structure of sememes, and we regard the 
sememes of all senses of a word together as the sememe set of the word. 

\begin{CJK*}{UTF8}{gbsn}
We now introduce the notation used in this paper. Let $G = (W, S, T)$ denotes 
the sememe KB, where $W = \{ w_1, w_2, \dots, w_{|W|} \}$ is the set of words, 
$S$ is the set of sememes, and $T \subseteq W \times S$ is the set of relation 
pairs between words and sememes. We denote the Chinese character set as $C$, 
with each word $w_i \in C^+$. Each word $w$ has its sememe set 
$S_w = \{ s | (w, s) \in T \}$. Take the word ``铁匠'' (ironsmith) for example, 
the sememe set $S_{\text{\it 铁匠\ (ironsmith)}}$ consists of ``人'' (human), 
``职位'' (occupation), ``金属'' (metal) and ``工'' (industrial).
\end{CJK*}

Given a word $w \in C^+$, the task of lexical sememe prediction aims to predict 
the corresponding $P(s | w)$ of sememes in $S$ to recommend them to $w$.

\section{Methodology}
In this section, we present our framework for lexical sememe prediction (SP). 
For each unlabelled word, our framework aims to recommend the most appropriate 
sememes based on the internal and external information. Because of introducing 
character information, our framework can work for both high-frequency and 
low-frequency words.

Our framework is the ensemble of two parts: sememe prediction with internal 
information (i.e., {\it internal} models), and sememe prediction with external 
information (i.e., {\it external} models). Explicitly, we adopt SPWE, SPSE, and 
their ensemble \cite{xie2017lexical} as {\it external} models, and we take 
SPWCF, SPCSE, and their ensemble as {\it internal} models.

In the following sections, we first introduce SPWE and SPSE. Then, we show the 
details of SPWCF and SPCSE. Finally, we present the method of model ensembling.

\subsection{SP with External Information}
\label{external}
SPWE and SPSE are introduced by \newcite{xie2017lexical} as the state of the art
for sememe prediction. These methods represent word meanings with embeddings 
learned from external information, and apply the ideas of collaborative 
filtering and matrix factorization in recommendation systems for sememe 
predication.

\textbf{SP with Word Embeddings (SPWE)} is based on the assumption that similar 
words should have similar sememes. In SPWE, the similarity of words are measured 
by cosine similarity. The score function $P(s_j | w)$ of sememe $s_j$ given a 
word $w$ is defined as: 
\begin{equation}
  P(s_j | w) \sim \sum_{w_i \in W} \cos(\mathbf{w}, \mathbf{w_i})\cdot \mathbf{M}_{ij}\cdot c^{r_i},
  \label{spwe}
\end{equation}
where $\mathbf{w}$ and $\mathbf{w_i}$ are pre-trained word embeddings of words 
$w$ and $w_i$. $\mathbf{M}_{ij} \in {\{0, 1\}}$ indicates the annotation of 
sememe $s_j$ on word $w_i$, where $\mathbf{M}_{ij} = 1$ indicates the word 
$s_j \in S_{w_i}$ and otherwise is not. $r_i$ is the descend cosine word 
similarity rank between $w$ and $w_i$, and $c \in (0, 1)$ is a hyper-parameter.

\textbf{SP with Sememe Embeddings (SPSE)} aims to map sememes into the same 
low-dimensional space of the word embeddings to predict the semantic 
correlations of the sememes and the words. This method learns two embeddings
$\mathbf{s}$ and $\mathbf{\bar{s}}$ for each sememe by solving matrix 
factorization with the loss function defined as:
\begin{equation}
  \small
  \begin{split}
    \mathcal{L} & = \sum_{w_i \in W, s_j \in S} \left( \mathbf{w}_i \cdot \left(\mathbf{s}_j + \mathbf{\bar{s}}_j \right) + \mathbf{b}_i + \mathbf{b}_j' - \mathbf{M}_{ij}\right)^2 \\
                & + \lambda \sum_{s_j,s_k\in S} \left( \mathbf{s}_j \cdot \mathbf{\bar{s}}_k - \mathbf{C}_{jk} \right)^2,
  \end{split}
  \label{spse}
\end{equation}
where $\mathbf{M}$ is the same matrix used in SPWE. $\mathbf{C}$ indicates the 
correlations between sememes, in which $\mathbf{C}_{jk}$ is defined as the
point-wise mutual information $\text{PMI}(s_j, s_k)$. The sememe embeddings are 
learned by factorizing the word-sememe matrix $\mathbf{M}$  and the 
sememe-sememe matrix $\mathbf{C}$ synchronously with fixed word embeddings. 
$\mathbf{b}_i$ and $\mathbf{b}_j'$ denote the bias of $w_i$ and $s_j$, and 
$\lambda$ is a hyper-parameter. Finally, the score of sememe $s_j$ given a word 
$w$ is defined as:
\begin{equation}
  P(s_j | w) \sim \mathbf{w} \cdot \left(\mathbf{s}_j + \mathbf{\bar{s}}_j \right).
\end{equation}

\subsection{SP with Internal Information}
We design two methods for sememe prediction with only internal character 
information without considering contexts as well as pre-trained word embeddings.

\subsubsection{SP with Word-to-Character Filtering (SPWCF)}
Inspired by collaborative filtering \cite{sarwar2001item}, we propose to 
recommend sememes for an unlabelled word according to its similar words based on 
internal information. Instead of using pre-trained word embeddings, we consider
words as {\it similar} if they contain the same characters at the same 
positions.

\begin{figure}[htbp]
  \centering
  \includegraphics[scale=0.45]{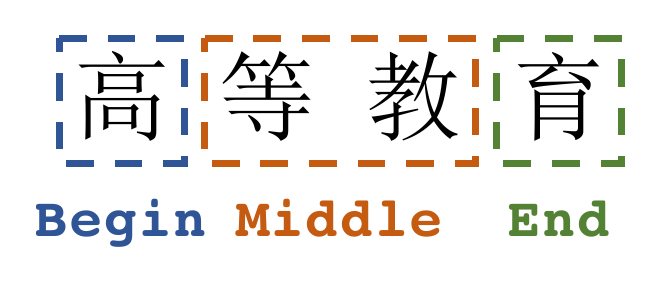}
  \caption{An example of the position of characters in a word.} 
  \label{fig:ins-railway-station}
\end{figure}

\begin{CJK*}{UTF8}{gbsn} 
In Chinese, the meaning of a character may vary according to its position 
within a word~\cite{chen2015joint}. We consider three positions within a word:
\texttt{Begin}, \texttt{Middle}, and \texttt{End}. For example, as shown in Fig.\ \ref{fig:ins-railway-station}, the character at the \texttt{Begin} position of 
the word ``火车站'' (railway station) is ``火'' (fire), while ``车'' (vehicle) and 
``站'' (station) are at the \texttt{Middle} and \texttt{End} position 
respectively. The character ``站'' usually means {\it station} when it is at the 
\texttt{End} position, while it usually means {\it stand} at the \texttt{Begin} 
position like in ``站立'' (stand), ``站岗哨兵'' (standing guard) and ``站起来'' 
(stand up).
\end{CJK*}

Formally, for a word $w = c_1c_2...c_{|w|}$, we define $\pi_{B}(w) = \{c_1\}$, 
$\pi_{M}(w) = \{c_2,...,c_{|w-1|}\}$, $\pi_{E}(w) = \{c_{|w|}\}$, and
\begin{equation}
  P_p(s_j | c) \sim \frac{\sum_{w_i \in W \land c \in \pi_{p}(w_i)}\mathbf{M}_{ij}}{\sum_{w_i \in W \land c \in \pi_{p}(w_i)} |S_{w_i}| },
\end{equation}
that represents the score of a sememe $s_j$ given a character $c$ and a position 
$p$, where $\pi_p$ may be $\pi_{B}$, $\pi_{M}$, or $\pi_{E}$. $\mathbf{M}$ is 
the same matrix used in Eq.\ (\ref{spwe}). Finally, we define the score function 
$P(s_j | w)$ of sememe $s_j$ given a word $w$ as:
\begin{equation}
  P(s_j | w) \sim \sum_{p \in \{B, M, E\}}\sum_{c \in \pi_{p}(w)} P_p(s_j | c).
\end{equation}

SPWCF is a simple and efficient method. It performs well because   
compositional semantics are pervasive in Chinese compound words, which makes it 
straightforward and effective to find similar words according to common 
characters.

\subsubsection{SP with Character and Sememe Embeddings (SPCSE)}
The method Sememe Prediction with Word-to-Character Filtering (SPWCF) can 
effectively recommend the sememes that have strong correlations with characters. 
However, just like SPWE, it ignores the relations between sememes. Hence, 
inspired by SPSE, we propose Sememe Prediction with Character and Sememe 
Embeddings (SPCSE) to take the relations between sememes into account. In SPCSE, 
we instead learn the sememe embeddings based on internal character information, 
then compute the semantic distance between sememes and words for prediction.

Inspired by GloVe \cite{pennington2014glove} and SPSE, we adopt matrix 
factorization in SPCSE, by decomposing the word-sememe matrix and the 
sememe-sememe matrix simultaneously. Instead of using pre-trained word 
embeddings in SPSE, we use pre-trained character embeddings in SPCSE. Since the 
ambiguity of characters is stronger than that of words, multiple embeddings are 
learned for each character \cite{chen2015joint}. We select the most 
representative character and its embedding to represent the word meaning. 
Because low-frequency characters are much rare than those low-frequency words, 
and even low-frequency words are usually composed of common characters, it is 
feasible to use pre-trained character embeddings to represent rare words. During 
factorizing the word-sememe matrix, the character embeddings are fixed. 

\begin{CJK*}{UTF8}{gbsn}
\begin{figure}[htbp]
  \centering
  \includegraphics[scale=0.35]{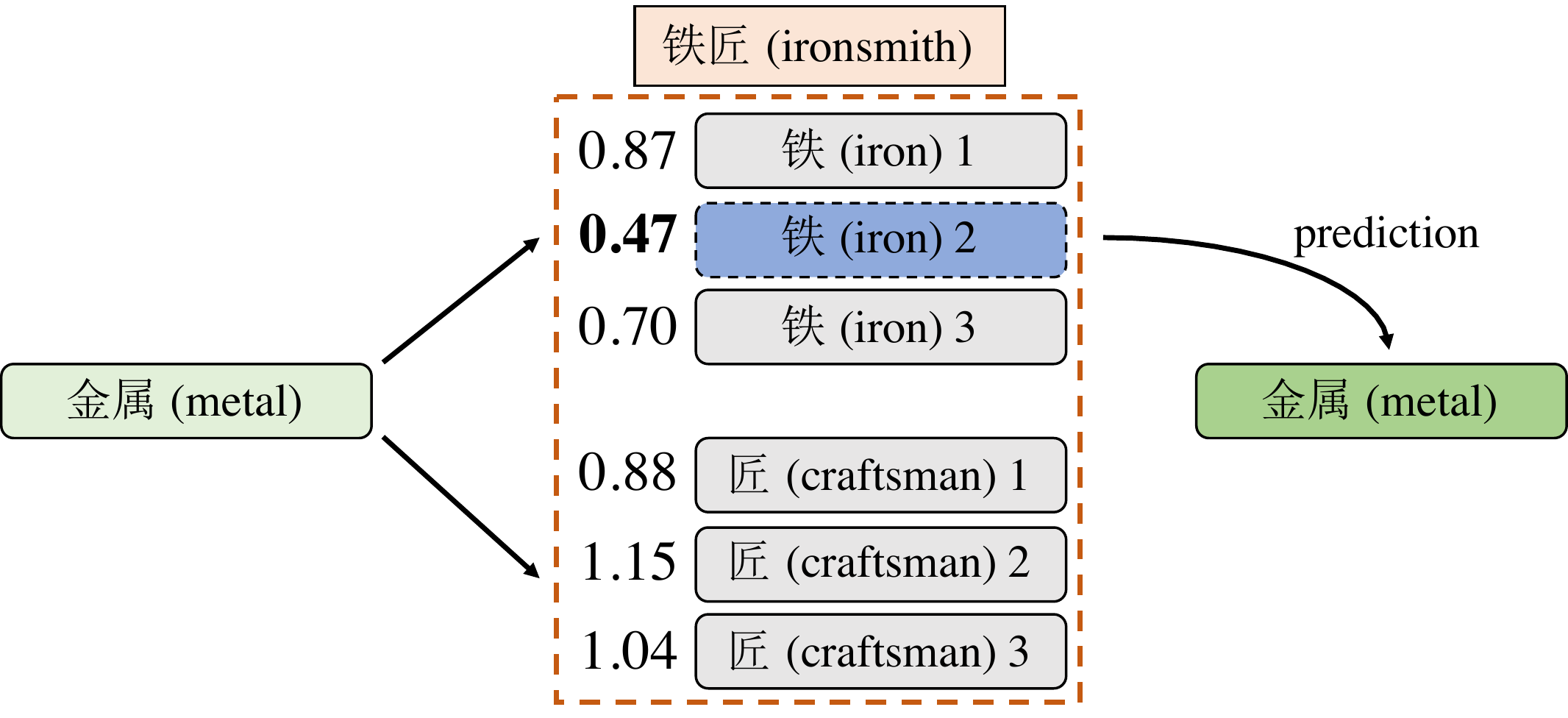}
  \caption{An example of adopting multiple-prototype character embeddings. The
            numbers are the cosine distances. The sememe ``金属'' (metal) is the 
            closest to one embedding of ``铁'' (iron).} 
  \label{fig:ins-cwe-blacksmith}
\end{figure}
\end{CJK*} 

We set $N_e$ as the number of embeddings for each character, and each character 
$c$ has $N_e$ embeddings $\mathbf{c}^1,...,\mathbf{c}^{N_e}$. Given a word $w$ 
and a sememe $s$, we select the embedding of a character of $w$ closest to the 
sememe embedding by cosine distance as the representation of the word $w$, as 
shown in Fig.\ \ref{fig:ins-cwe-blacksmith}. Specifically, given a word 
$w=c_1...c_{|w|}$ and a sememe $s_j$, we define
\begin{equation}
  \hat{k}, \hat{r} =\arg\min_{k, r}\left[ 1 - \cos( \mathbf{c}_k^{r} , (\mathbf{s}'_j+\mathbf{\bar{s}}_j'))\right],
  \label{eq:rkmax}
\end{equation}
where $\hat{k}$ and $\hat{r}$ indicate the indices of the character and its 
embedding closest to the sememe $s_j$ in the semantic space. With the same 
word-sememe matrix $\mathbf{M}$ and sememe-sememe correlation matrix 
$\mathbf{C}$ in Eq.\ (\ref{spse}), we learn the sememe embeddings with the loss 
function:
\begin{equation}
\small
\begin{split}
\mathcal{L} & = \sum_{w_i \in W, s_j \in S} \left( \mathbf{c}_{\hat{k}}^{\hat{r}} \cdot \left(\mathbf{s}_j' + \mathbf{\bar{s}}_j' \right) + \mathbf{b}_{\hat{k}}^c + \mathbf{b}_j'' - \mathbf{M}_{ij}\right)^2 \\
            & + \lambda' \sum_{s_j,s_q\in S} \left( \mathbf{s}_j' \cdot \mathbf{\bar{s}}_q' - \mathbf{C}_{jq} \right)^2,
\end{split}
\label{eq:spcse}
\end{equation}
where $\mathbf{s}_j'$ and $\mathbf{\bar{s}}_j'$ are the sememe embeddings for 
sememe $s_j$, and $\mathbf{c}_{\hat{k}}^{\hat{r}}$ is the embedding of the 
character that is the closest to sememe $s_j$ within $w_i$. Note that, as the 
characters and the words are not embedded into the same semantic space, we learn 
new sememe embeddings instead of using those learned in SPSE, hence we use 
different notations for the sake of distinction. $\mathbf{b}_{k}^c$ and 
$\mathbf{b}_j''$ denote the biases of $c_k$ and $s_j$, and $\lambda'$ is the 
hyper-parameter adjusting the two parts. Finally, the score function of word 
$w=c_1...c_{|w|}$ is defined as:
\begin{equation}
  \begin{split}
  P(s_j | w) \sim \mathbf{c}_{\hat{k}}^{\hat{r}} \cdot \left(\mathbf{s}_j' + \mathbf{\bar{s}}_j' \right).
  \end{split}
  \label{eq:spcse-score}
\end{equation}

\subsection{Model Ensembling}
SPWCF / SPCSE and SPWE / SPSE take different sources of information as input, 
which means that they have different characteristics: SPWCF / SPCSE only have 
access to internal information, while SPWE / SPSE can only make use of external 
information. On the other hand, just like the difference between SPWE and SPSE, 
SPWCF originates from collaborative filtering, whereas SPCSE uses matrix 
factorization. All of those methods have in common that they tend to recommend 
the sememes of {\it similar} words, but they diverge in their interpretation of 
{\it similar}.  
% SPWCF and SPCSE focus on internal similarity, including the morphological 
% information and the contexts of characters. Meanwhile, SPWE and SPSE both take 
% the advantage of external similarity. Hence if two words usually appear in 
% similar contexts, then SPWE and SPSE will assume that they share the similar 
% sememes, but some self-explanatory semantic meanings may not always show in 
% contexts.

\begin{figure}[htbp]
  \centering
  \includegraphics[scale=0.25]{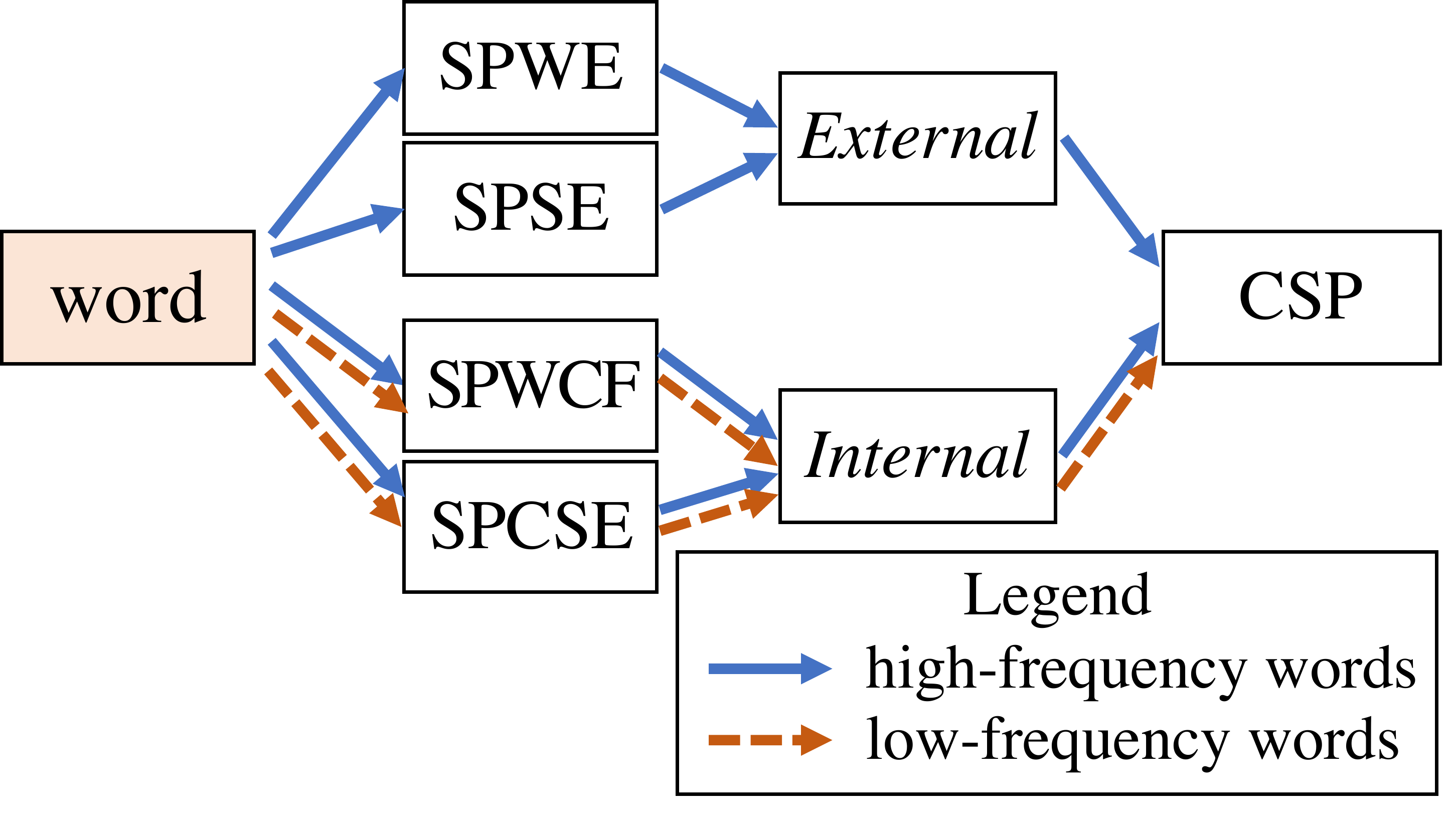}
  \caption{The illustration of model ensembling.} 
  \label{fig:illus-model-ensemble}
\end{figure}

Hence, to obtain better prediction performance, it is necessary to combine  
these models. We denote the ensemble of SPWCF and SPCSE as the
{\it internal} model, and we denote the ensemble of SPWE and SPSE as the 
{\it external} model. The ensemble of the {\it internal} and the {\it external} 
models is our novel framework CSP. In practice, for words with reliable word 
embeddings, i.e., high-frequency words, we can use the integration of the 
{\it internal} and the {\it external} models; for words with extremely low 
frequencies (e.g., having no reliable word embeddings), we can just use the 
{\it internal} model and ignore the {\it external} model, because the external 
information is noise in this case. Fig.~\ref{fig:illus-model-ensemble} shows 
model ensembling in different scenarios. For the sake of comparison, we use the 
integration of SPWCF, SPCSE, SPWE, and SPSE as CSP in our all experiments. In 
this paper, two models are integrated by simple weighted addition. 

\section{Experiments} 
In this section, we evaluate our models on the task of sememe 
prediction. Additionally, we analyze the performance of different methods for 
various word frequencies. We also execute an elaborate case study to 
demonstrate the mechanism of our methods and the advantages of using internal 
information.

\subsection{Dataset}
We use the human-annotated sememe KB HowNet for sememe prediction. In HowNet, 
$103,843$ words are annotated with $212,539$ senses, and each sense is defined 
as a hierarchical structure of sememes. There are about $2,000$ sememes in 
HowNet. However, the frequencies of some sememes in HowNet are very low, such 
that we consider them unimportant and remove them. 
%As a result, we keep 
%$1,400$ sememes in our dataset. 
Our final dataset contains $1,400$ sememes. For learning the word and character 
embeddings, we use the Sogou-T corpus\footnote{Sogou-T corpus is provided by 
Sogou Inc., a Chinese commercial search engine company.
\url{https://www.sogou.com/labs/resource/t.php}} \cite{liu2012identifying}, 
which contains $2.7$ billion words.

\subsection{Experimental Settings}
In our experiments, 
we evaluate SPWCF, SPCSE, and SPWCF + SPCSE which only use internal information, 
and the ensemble framework CSP which uses both internal and external information 
for sememe prediction. We use the state-of-the-art models from 
\newcite{xie2017lexical} as our baselines. Additionally, we use the SPWE model 
with word embeddings learned by fastText \cite{bojanowski2017enriching} that 
considers both internal and external information as a baseline.

For the convenience of comparison, we select $60,000$ high-frequency words in 
Sogou-T corpus from HowNet. We divide the $60,000$ words into train, dev, and 
test sets of size $48,000$, $6,000$, and $6,000$, respectively, and we keep them 
fixed throughout all experiments except for Section \ref{sec:frequency}. In 
Section \ref{sec:frequency}, we utilize the same train and dev sets, but use 
other words from HowNet as the test set to analyze the performance of our 
methods for different word frequency scenarios. We select the hyper-parameters 
on the dev set for all models including the baselines and report the evaluation 
results on the test set.

We set the dimensions of the word, sememe, and character embeddings to be $200$. 
The word embeddings are learned by GloVe \cite{pennington2014glove}.
For the baselines, in SPWE, the hyper-parameter $c$ is set to $0.8$, and the 
model considers no more than $K=100$ nearest words. We set the probability of 
decomposing zero elements in the word-sememe matrix in SPSE to be $0.5\%$. 
$\lambda$ in Eq.\ (\ref{spse}) is $0.5$. The model is trained for $20$ epochs, 
and the initial learning rate is $0.01$, which decreases through iterations. 
For fastText, we use skip-gram with hierarchical softmax to learn word 
embeddings, and we set the minimum length of character n-grams to be 1 and the 
maximum length of character n-grams to be 2. For model ensembling, we use 
$\frac{\lambda_{\text{\it SPWE}}}{\lambda_{\text{\it SPSE}}} = 2.1$ as the 
addition weight.

For SPCSE, we use Cluster-based Character Embeddings \cite{chen2015joint} to
learn pre-trained character embeddings, and we set $N_e$ to be $3$. We set 
$\lambda'$ in Eq.\ (\ref{eq:spcse}) to be $0.1$, and the model is trained for 
$20$ epochs. The initial learning rate is $0.01$ and decreases during training 
as well. Since generally each character can relate to about $15$ - $20$ sememes, 
we set the probability of decomposing zero elements in the word-sememe matrix in 
SPCSE to be $2.5\%$. The ensemble weight of SPWCF and SPCSE 
$\frac{\lambda_{\text{\it SPWCF}}}{\lambda_{\text{\it SPCSE}}} = 4.0$. For 
better performance of the final ensemble model CSP, we set $\lambda = 0.1$ and 
$\frac{\lambda_{\text{\it SPWE}}}{\lambda_{\text{\it SPSE}}} = 0.3125$, though 
$0.5$ and $2.1$ are the best for SPSE and SPWE + SPSE. Finally, we choose 
$\frac{\lambda_{\text{\it internal}}}{\lambda_{\text{\it external}}} = 1.0$ to 
integrate the {\it internal} and {\it external} models.

\subsection{Sememe Prediction}

\subsubsection{Evaluation Protocol}
The task of sememe prediction aims to recommend appropriate sememes for 
unlabelled words. We cast this as a multi-label classification task, and adopt 
mean average precision (MAP) as the evaluation metric. For each unlabelled word 
in the test set, we rank all sememe candidates with the scores given by our 
models as well as baselines, and we report the MAP results. The results are 
reported on the test set, and the hyper-parameters are tuned on the dev set.

\subsubsection{Experiment Results}
The evaluation results are shown in Table \ref{tab:result-sememe-prediction-high}. 
%From the evaluation results,
We can observe that:

\begin{table}[htbp]
  \centering
  \small
  \resizebox{\linewidth}{!}{
  \begin{tabular}{p{0.5cm}lp{0.25cm}cp{0.5cm}}
    \toprule
        &  Method             &   & MAP         & \\
    \midrule
        &  SPSE               &   & 0.411       & \\
        &  SPWE               &   & 0.565       & \\
        &  SPWE+SPSE          &   & 0.577       & \\
    \midrule
        &  SPWCF              &   & 0.467       & \\
        &  SPCSE              &   & 0.331       & \\
        &  SPWCF + SPCSE      &   & {\bf 0.483} & \\
    \midrule
        &  SPWE + fastText    &   & 0.531       & \\
        &  CSP                &   & {\bf 0.654} & \\
    \bottomrule
  \end{tabular}}
  \caption{\label{tab:result-sememe-prediction-high} Evaluation results on sememe 
          prediction. The result of SPWCF + SPCSE is bold for comparing with other 
          methods (SPWCF and SPCSE) which use only internal information.}
\end{table}

%\begin{table*}
%  \centering
%  \small
%  \begin{tabular}{ccccccccc}
%    \toprule
%      word frequency  & occurrences & SPWE  & SPSE  & SPWE + SPSE & SPWCF   & SPCSE & SPWCF + SPCSE & CSP   \\
%    \midrule
%      $<$ 10          & 1,755		        & 0.195 & 0.113 & 0.168       & 0.448 & 0.306	& 0.459       & {\bf 0.471} \\	
%      10 - 50         & 6,782		        & 0.342 & 0.206 & 0.314	      & 0.458 & 0.310	& 0.469       & {\bf 0.542} \\	
%      50 - 100        & 4,897           & 0.437 & 0.273 & 0.414       & 0.414 & 0.291	& 0.437       & {\bf 0.555} \\	    
%      100 - 1000      & 5,344		        & 0.481 & 0.339 & 0.478       & 0.400	& 0.286	& 0.418       & {\bf 0.555} \\
%      1000 - 5000     & 3,855           & 0.558 & 0.409 & 0.556	      & 0.443	& 0.312	& 0.456       & {\bf 0.626}	\\
%      5000 - 10000    & 1,318           & 0.549	& 0.407 & 0.548       & 0.462	& 0.339	& 0.477       & {\bf 0.632} \\	
%      10000 - 30000   & 1,457           & 0.556 & 0.424 & 0.554       & 0.463	& 0.353	& 0.477       & {\bf 0.641} \\	
%      $>$ 30000       & 1,371           & 0.509 & 0.386 & 0.511	      & 0.479	& 0.342	& 0.494       & {\bf 0.624} \\	
%    \bottomrule
%  \end{tabular}
%  \caption{\label{tab:sememe-frequencies} MAP scores on sememe prediction with different word frequencies.}
%\end{table*}

\begin{table*}
  \centering
  \small
  \begin{tabular}{c|cccccccc}
    \toprule
      word frequency  & $\leqslant$ 50  & 51-- 100    & 101 -- 1,000& 1,001 -- 5,000 & 5,001 -- 10,000& 10,001 -- 30,000& $>$30,000   \\
      occurrences     & 8537            & 4868        & 3236        & 2036          & 663             & 753             & 686         \\ 
    \midrule
      SPWE            & 0.312           & 0.437       & 0.481       & 0.558         & 0.549           & 0.556           & 0.509	      \\
      SPSE            & 0.187           & 0.273       & 0.339       & 0.409         & 0.407           & 0.424	          & 0.386	      \\
      SPWE + SPSE     & 0.284           & 0.414       & 0.478	      & 0.556	        & 0.548	          & 0.554	          & 0.511       \\
      SPWCF           & 0.456           & 0.414	      & 0.400	      & 0.443	        & 0.462	          & 0.463	          & 0.479       \\
      SPCSE           & 0.309           & 0.291	      & 0.286	      & 0.312	        & 0.339	          & 0.353	          & 0.342       \\
      SPWCF + SPCSE   & 0.467           & 0.437	      & 0.418	      & 0.456	        & 0.477	          & 0.477	          & 0.494       \\
      SPWE + fastText & 0.495           & 0.472	      & 0.462	      & 0.520	        & 0.508	          & 0.499	          & 0.490       \\
      CSP             & {\bf 0.527}     & {\bf0.555}  & {\bf 0.555}	& {\bf 0.626}   & {\bf 0.632}	    & {\bf 0.641}	    & {\bf 0.624} \\			
    \bottomrule
  \end{tabular}
  \caption{\label{tab:sememe-frequencies} MAP scores on sememe prediction with different word frequencies.}
\end{table*}
 
(1) Considerable improvements are obtained via model ensembling, and the CSP 
model achieves state-of-the-art performance. CSP combines the internal character 
information with the external context information, which significantly and 
consistently improves performance on sememe prediction. Our results confirm the 
effectiveness of a combination of internal and external information for sememe 
prediction; since different models focus on different features of the inputs, 
the ensemble model can absorb the advantages of both methods. 

(2) The performance of SPWCF + SPCSE is better than that of SPSE, which means 
using only internal information could already give good results for sememe 
prediction as well. Moreover, in {\it internal} models, SPWCF performs much 
better than SPCSE, which also implies the strong power of collaborative 
filtering.

(3) The performance of SPWCF + SPCSE is worse than SPWE + SPSE. This indicates 
that it is still difficult to figure out the semantic meanings of a word without 
contextual information, due to the ambiguity and meaning vagueness of internal 
characters. Moreover, some words are not compound words (e.g., single-morpheme 
words or transliterated words), whose meanings can hardly be inferred directly 
by their characters. In Chinese, internal character information is just partial 
knowledge. We present the results of SPWCF and SPCSE merely to show the 
capability to use the internal information in isolation. In our case study, we 
will demonstrate that {\it internal} models are powerful for low-frequency words, and 
can be used to predict senses that do not appear in the corpus. 
% It is significant in most cases to combine the internal information and 
% external information. 

\subsection{Analysis on Different Word Frequencies}
\label{sec:frequency}

To verify the effectiveness of our models on different word frequencies, we 
incorporate the remaining words in HowNet\footnote{In detail, we do not use the 
numeral words, punctuations, single-character words, the words do not appear in 
Sogou-T corpus (because they need to appear at least for one time to get the 
word embeddings), and foreign abbreviations.} into the test set. Since the 
remaining words are low-frequency, we mainly focus on words with long-tail 
distribution. We count the number of occurrences in the corpus for each word in 
the test set and group them into eight categories by their frequency. The 
evaluation results are shown in Table \ref{tab:sememe-frequencies}, from which 
we can observe that:

\begin{CJK*}{UTF8}{gbsn}
  \begin{table*}
    \centering
    \small
    \resizebox{\linewidth}{!}{
    \begin{tabular}{ccl}
      \toprule
        words                     & models    & \multicolumn{1}{c}{Top 5 sememes} \\
      \hline
        \multirow{3}{*}{\shortstack{钟表匠\\(clockmaker)}}   & internal  & {\bf 人(human)}, {\bf 职位(occupation)}, 部件(part), {\bf 时间(time)},  {\bf 告诉(tell)} \\
                                                            & external  & {\bf 人(human)}, 专(ProperName), 地方(place), 欧洲(Europe), 政(politics)\\ 
                                                            & ensemble  & {\bf 人(human)}, {\bf 职位(occupation)}, {\bf 告诉(tell)}, {\bf 时间(time)}, {\bf 用具(tool)} \\
      \hline
        \multirow{3}{*}{\shortstack{奥斯卡\\(Oscar)}}        & internal  & {\bf 专(ProperName)}, 地方(place), 市(city), 人(human), 国都(capital) \\ 
                                                            & external  & {\bf 奖励(reward)}, {\bf 艺(entertainment)}, {\bf 专(ProperName)}, 用具(tool), {\bf 事情(fact)} \\
                                                            & ensemble  & {\bf 专(ProperName)}, {\bf 奖励(reward)}, {\bf 艺(entertainment)}, 著名(famous), 地方(place) \\
      \bottomrule
    \end{tabular}}
    \caption{Examples of sememe prediction. For each word, we present the top $5$ sememes predicted by the {\it internal} model, {\it external} model and the final ensemble model (CSP). Bold sememes are correct. \label{tab:sememe-case-study}}
  \end{table*} 
\end{CJK*}

(1) The performances of SPSE, SPWE, and SPWE + SPSE decrease dramatically with 
low-frequency words compared to those with high-frequency words. On the 
contrary, the performances of SPWCF, SPCSE, and SPWCF~+~SPCSE, though weaker 
than that on high-frequency words, is not strongly influenced in the long-tail 
scenario. The performance of CSP also drops since CSP also uses external 
information, which is not sufficient with low-frequency words. These results 
show that the word frequencies and the quality of word embeddings can influence 
the performance of sememe prediction methods, especially for {\it external} models 
which mainly concentrate on the word itself. However, the {\it internal} models are 
more robust when encountering long-tail distributions. Although words 
do not need to appear too many times for learning good word embeddings, it is 
still hard for {\it external models} to recommend sememes for low-frequency words. 
While since {\it internal} models do not use external word embeddings, they can still 
work in such scenario. As for the performance on high-frequency words, since 
these words are used widely, the ambiguity of high-frequency words is thus much 
stronger, while the {\it internal} models are still stable for high-frequency words.

(2) The results also indicate that even low-frequency words in Chinese are 
mostly composed of common characters, and thus it is possible to utilize 
internal character information for sememe prediction on words with long-tail 
distribution (even on those new words that never appear in the corpus). 
Moreover, the stability of the MAP scores given by our methods on various word 
frequencies also reflects the reliability and universality of our models in 
real-world sememe annotations in HowNet. We will give detailed analysis in our 
case study.

\subsection{Case Study}
The results of our main experiments already show the effectiveness of our 
models. In this case study, we further investigate the outputs of our models to 
confirm that character-level knowledge is truly incorporated into sememe 
prediction. 
% Meanwhile, we want to confirm the mechanism behind our models.

%\subsubsection{Analysis on Predicted Sememes}

\begin{CJK*}{UTF8}{gbsn}
In Table \ref{tab:sememe-case-study}, we demonstrate the top $5$ sememes 
for ``钟表匠'' (clockmaker) and ``奥斯卡'' (Oscar, i.e., the Academy Awards). 
``钟表匠'' (clockmaker) is a typical compound word, while ``奥斯卡'' (Oscar) is a 
transliterated word. For each word, the top $5$ results generated by the 
internal model (SPWCF + SPCSE), the {\it external} model (SPWE + SPSE) and 
the ensemble model (CSP) are listed.
 
The word ``钟表匠'' (clockmaker) is composed of three characters: ``钟'' (bell, 
clock), ``表'' (clock, watch) and ``匠'' (craftsman). Humans can intuitively 
conclude that {\it clock} $+$ {\it craftsman} $\,\to\,$ {\it clockmaker}. 
However, the {\it external} model does not perform well for this example. If we 
investigate the word embedding of ``钟表匠'' (clockmaker), we can know why this 
method recommends these unreasonable sememes. 
The closest $5$ words in the train set to 
``钟表匠'' (clockmaker) by cosine similarity of their embeddings 
are: ``瑞士'' (Switzerland), ``卢梭'' (Jean-Jacques Rousseau), ``鞋匠'' (cobbler), 
``发明家'' (inventor) and ``奥地利人'' (Austrian). Note that none of these words 
are directly relevant to {\it bells}, {\it clocks} or {\it watches}. Hence, the 
sememes ``时间'' (time), ``告诉'' (tell), and ``用具'' (tool) cannot be inferred 
by those words, even though the correlations between sememes are introduced by 
SPSE. In fact, those words are related to {\it clocks} in an indirect way: 
Switzerland is famous for watch industry; Rousseau was born into a family that 
had a tradition of watchmaking; cobbler and inventor are two kinds of 
occupations as well. With the above reasons, those words usually co-occur with 
``钟表匠'' (clockmaker), or usually appear in similar contexts as ``钟表匠'' 
(clockmaker). It indicates that related word embeddings as used in an 
{\it external} model do not always recommend related sememes.

The word ``奥斯卡'' (Oscar) is created by the pronunciation of {\it Oscar}. 
Therefore, the meaning of each character in ``奥斯卡'' (Oscar) is unrelated to 
the meaning of the word. Moreover, the characters ``奥'', ``斯'', and ``卡'' are 
common among transliterated words, thus the {\it internal} method recommends 
``专'' (ProperName) and ``地方'' (place), etc., since many transliterated words 
are proper nouns or place names.
\end{CJK*}

\section{Conclusion and Future Work}
In this paper, we introduced character-level internal information for lexical 
sememe prediction in Chinese, in order to alleviate the problems caused by the 
exclusive use of external information. We proposed a Character-enhanced Sememe 
Prediction (CSP) framework which integrates both internal and external 
information for lexical sememe prediction and proposed two methods for utilizing 
internal information. We evaluated our CSP framework on the classical manually 
annotated sememe KB HowNet. In our experiments, our methods achieved promising 
results and outperformed the state of the art on sememe prediction, especially 
for low-frequency words.

We will explore the following research directions in the future: (1) Concepts in 
HowNet are annotated with hierarchical structures of senses and sememes, but 
those are not considered in this paper. In the future, we will take structured 
annotations into account. (2) It would be meaningful to take more information
into account for blending external and internal information and design more 
sophisticated methods. (3) Besides Chinese, many other languages have rich 
subword-level information. In the future, we will explore methods of exploiting 
internal information in other languages. (4) We believe that sememes are 
universal for all human languages. We will explore a general framework to 
recommend and utilize sememes for other NLP tasks. 
  
\section*{Acknowledgments}
This research is part of the NExT++ project, supported by the National Research 
Foundation, Prime Minister's Office, Singapore under its IRC@Singapore Funding 
Initiative. This work is also supported by the National Natural Science 
Foundation of China (NSFC No. 61661146007 and 61572273) and the research fund of
Tsinghua University-Tencent Joint Laboratory for Internet Innovation Technology. 
Hao Zhu is supported by Tsinghua University Initiative Scientific Research 
Program. We would like to thank Katharina Kann, Shen Jin, and the anonymous 
reviewers for their helpful comments.
\vspace{1em}
%\bibliography{acl2018}
\bibliography{acl2018}
\bibliographystyle{acl_natbib}

\appendix

\end{document}